\documentclass[fleqn,10pt]{wlscirep}
\usepackage[utf8]{inputenc}
\usepackage[T1]{fontenc}
\usepackage{xcolor}
\usepackage{lineno}
\usepackage{amsmath}
\usepackage{enumitem}
\usepackage{multirow} 
\usepackage{array} 
\usepackage{hyperref} 
\usepackage[perpage,symbol]{footmisc} 

\makeatletter
\renewcommand{\@fnsymbol}[1]{%
    \ifcase#1\or †\or *\or ‡\or §\or ¶\or **\or ††\or ‡‡\else\@ctrerr\fi}
\makeatother

\title{InterHub: A Naturalistic Trajectory Dataset with Dense Interaction for Autonomous Driving}

\author[1]{Xiyan Jiang}
\author[1,*]{Xiaocong Zhao}
\author[1]{Yiru Liu}
\author[2]{Zirui Li}
\author[1]{Peng Hang}
\author[3]{Lu Xiong}
\author[1,*]{Jian Sun}

\affil[1]{Key Laboratory of Road and Traffic Engineering, Ministry of Education, Tongji University, Shanghai, 201804, China}
\affil[2]{School of Mechanical Engineering, Beijing Institute of Technology, Beijing 100081, China}
\affil[3]{School of Automotive Studies, Tongji University, Shanghai, 201804, China}
\affil[*]{corresponding author: Xiaocong Zhao (zhaoxc@tongji.edu.cn), Jian Sun (sunjian@tongji.edu.cn)}

\begin{abstract}
The driving interaction—a critical yet complex aspect of daily driving—lies at the core of autonomous driving research. However, real-world driving scenarios sparsely capture rich interaction events, limiting the availability of comprehensive trajectory datasets for this purpose. To address this challenge, we present \texttt{InterHub}, a dense interaction dataset derived by mining interaction events from extensive naturalistic driving records. We employ formal methods to describe and extract multi-agent interaction events, exposing the limitations of existing autonomous driving solutions. Additionally, we introduce a user-friendly toolkit enabling the expansion of \texttt{InterHub} with both public and private data. By unifying, categorizing, and analyzing diverse interaction events, \texttt{InterHub} facilitates cross-comparative studies and large-scale research, thereby advancing the evaluation and development of autonomous driving technologies.

\end{abstract}

\begin{document}

\flushbottom
\maketitle

\thispagestyle{empty}


\section*{Background \& Summary}
The driving interaction is one of the most challenging parts of daily driving, gaining increasing attention in fields ranging from developing to validating the autonomous vehicle (AV). Involving complex negotiation, coordination, cooperation, and competition among vehicles, cyclists, and pedestrians, driving interactions can lead to traffic flow obstruction or even accidents if handled with misinterpretation of behavior or deviation from expected traffic norms \cite{rasouli2019autonomous, pelikan2021autonomous}. When interacting with human road users, current AV systems often prioritize safety over efficiency, resulting in overly cautious behaviors or failed interactions. Such shortcomings can erode public trust and hinder the acceptance of AVs \cite{koh2023public}.

To enhance AV performance in interactive scenarios, researchers emphasize the need for datasets containing dense driving interactions \cite{driggs2017integrating}. The driving interaction, referring to driving scenarios where at least two vehicles are influenced by one another in their near future, has become a focal point across various domains of autonomous driving, such as trajectory prediction \cite{zhang2020trajectory, min2020interaction, ding2019predicting}, decision-making \cite{chen2023interaction,huang2023learning,zhang2024interactive}, and motion-planning\cite{shu2023human, arbabi2022planning, li2019interaction, liu2022interaction, wang2023safety}. Researchers strive to extract typical human interaction behaviors from a diverse array of driving record collections, aiming to reveal and learn interacting strategies from comprehensive human demonstrations in intensive interactions, which are usually safety-critical. Great efforts have also been spread to validate AV's interacting ability by constructing interactive traffic flow simulation\cite{paz2015traffic, hasan2021distributed, yang2019real, chao2015vehicle}, and accelerated-validation-orientated testing environments \cite{liu2024towards, zhang2022human, wei2024interactive}. These works focus on creating realistic and challenging interaction scenarios by learning behavioral patterns from real-world driving records. 

However, the scarce nature of driving interaction events in the natural driving poses a challenge in accessing large-scale dedicated interaction datasets, leading to the absence of a unified data foundation for interaction-related studies.  In-depth studies often rely on purpose-specific datasets. For example, car-following datasets have been well-established in recent years and serving longitudinal driving behavior researches \cite{chen2023follownet, zhou2024a}. While existing naturalistic driving datasets, such as Lyft Level 5\cite{houston2021one}, Waymo Open Motion Dataset (WOMD)\cite{ettinger2021waymo}, and INTERACTION dataset\cite{zhan2019interaction}, offer substantial amounts of driving clips, capturing effective interaction scenes is particularly challenging due to its inherent spatial and temporal sparsity. Ivanovic\textit{ et al.}\cite{ivanovic2024trajdata} noted that a considerable percentage of agents in many datasets remain stationary, with stationary vehicles comprising up to 53.6\% in the WOMD. Furthermore, even moving vehicles do not always interact in simple tasks, such as free driving. Researchers utilize distinct approaches to identify driving interaction events\cite{arbabi2022planning, ozkan2021socially, liu2022interaction}, according to varied definitions of driving interactions\cite{svensson1998method, markkula2020defining, rasouli2019autonomous, madigan2019understanding}. However, existing definitions for the driving interaction usually do not guarantee a precise and quantitative description, leaving ambiguity in the spatial and temporal boundaries of extracted events. The absence of a standardized data benchmark hinders direct comparisons across studies, making it challenging to measure the overall advancements in the field of driving-interaction-related research. The limited size of datasets may even lead to unreliable research findings. The conclusions from most of the driving-interaction-related studies rely heavily on interaction clips which are usually from a single dataset. This dependence, coupled with the inherent limitations and lack of diversity in a single dataset, can lead to conclusions that may not be broadly applicable across different contexts. For instance, Li\textit{ et al.} \cite{li2023large} noted in their study that the constrained scope of the datasets usage led to conflicting conclusions\cite{wen2022characterizing, hu2023autonomous} regarding the behaviors patterns in human-only and human-autonomous interactions. 

\begin{figure}
    \centering
    \includegraphics[width=1\linewidth]{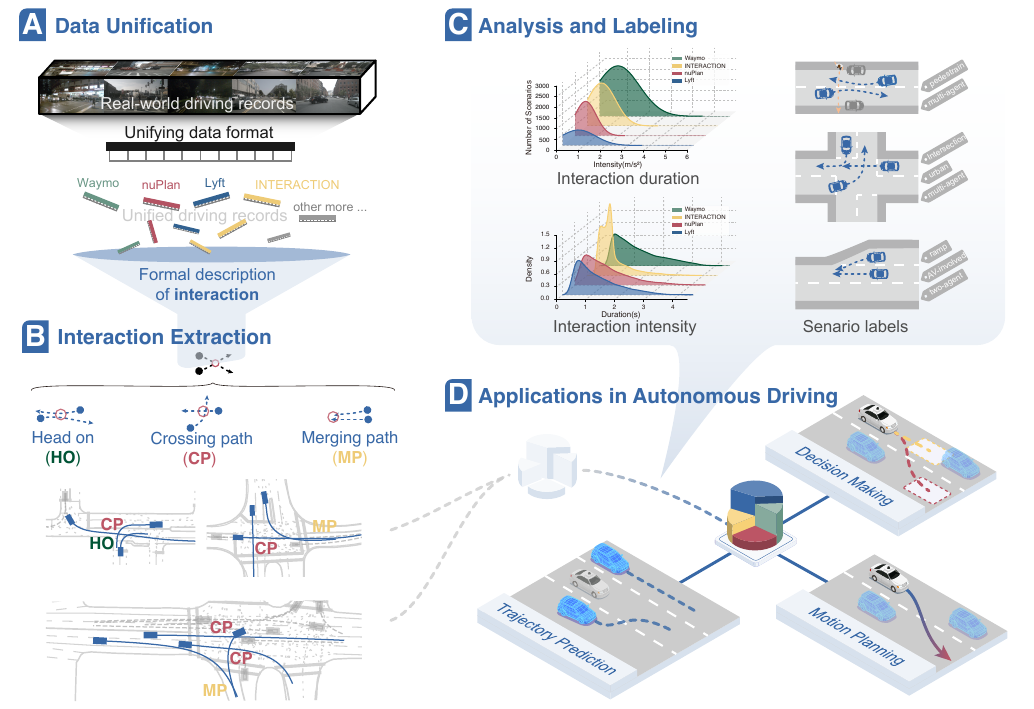}
    \caption{Road map of dense driving interaction dataset construction and application.}
    \label{fig:roadmap}
\end{figure}

Our work aims to provide a unified and extensive data foundation for driving interaction research by mining driving interaction events from public naturalistic driving datasets. We show the road map of our work in Figure \ref{fig:roadmap}. Firstly, widely adopted naturalistic driving datasets are reorganized using a unified data interface to provide extensibility and easy access to multiple driving data resources. Then, driving interaction events covering a wide range of interaction archetypes, as well as their combinations, are extracted using the formal method. Rich features of the extracted scenarios, including interaction intensity, AV involvement, and conflict type, are analyzed and annotated to support applications with varied needs regarding driving interaction data. 
Key contributions of this work are as follows:

\begin{itemize}
    
    \setlength{\itemsep}{0.001\baselineskip}
    \item A naturalistic driving dataset with dense interactions is extracted from multiple widely adopted public driving datasets, being unified, analyzed, and categorized to provide a user-friendly data foundation, \texttt{InterHub}, for driving-interaction-centered research.
    
    \setlength{\itemsep}{0.001\baselineskip}
    \item A quantitative-yet-readable definition of the driving interaction using the formal method is proposed along with an open access toolkit, serving cross-discipline applicable interaction description and allowing user-side extraction of multi-agent driving interaction events.
    
    \setlength{\itemsep}{0.001\baselineskip}
    \item Experiments are conducted to show the challenging scenarios in \texttt{InterHub} which is valuable for validating AV performance in key tasks including trajectory prediction and motion planning.
    
    \setlength{\itemsep}{0.001\baselineskip}
\end{itemize}

\begin{figure}
    \centering
    \includegraphics[width=1\linewidth]{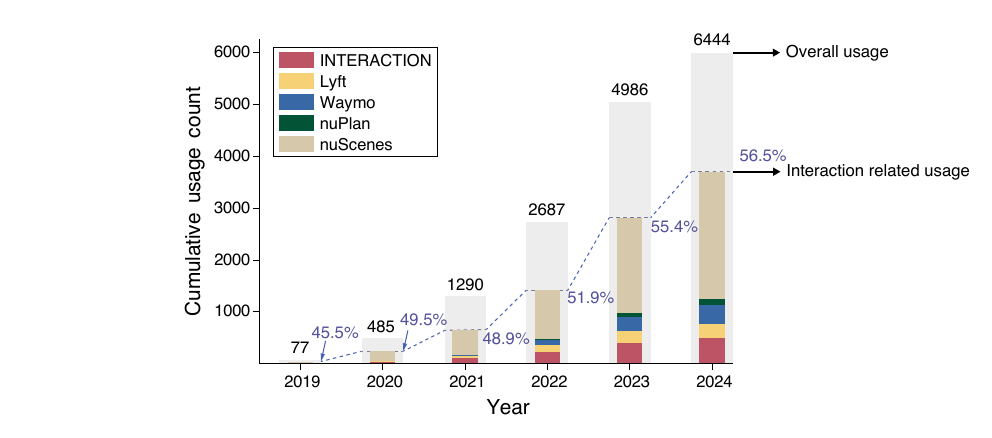}
    \caption{Accumulated usage of widely adopted naturalistic driving datasets.}
    \label{fig:usage-freq}
\end{figure}

\subsection*{Related works}

Recent years witnessed substantial exploration in the field of driving interaction where the public naturalistic driving datasets contribute a momentum. Figure \ref{fig:usage-freq} illustrates the cumulative usage of naturalistic driving datasets in recent years, highlighting their widespread adoption and applications, especially in the field of driving interaction studies\footnote{Usage statistics are based on Google Scholar data as of October 31, 2024.}. Notably, except for the INTERACTION dataset\cite{zhan2019interaction} which is specifically featured for driving interaction events, other comprehensive datasets are also intensively applied in the interaction-related study, indicating the overflowing needs for interaction data. Thus, efforts have been widely rolled out to extract interaction events from large-scale datasets.

\textit{Semantic definition of driving interaction}. To gain recognition for research advancement, it is essential to lay works on publicly accepted benchmarks with a broad audience. This is especially true for empirical studies that heavily rely on the data. To set a publicly accepted research basis, the definition of the driving interaction is a common starting point. 

Interaction research encompasses multiple research fields such as sociology, psychology, and linguistics, and different disciplines interpret interaction from varied perspectives. Limiting the scope within the field of autonomous driving, the definition of driving interaction could be dated back as early as 1998, when Svensson \textit{et al.}\cite{svensson1998method} defined interaction as a prerequisite for avoiding accidents in traffic events with the possibility of collisions, setting a major reference for subsequent studies. On this basis, Markkula \textit{et al. }\cite{markkula2020defining} integrates theoretical perspectives from different disciplines and defines road traffic interaction as the behavior of at least two road users can be explained as being influenced by the possibility that they both intend to occupy the same spatial area in the near future. 

More recent studies leverage the awareness of the technology developing trend and spare more attention to the needs of autonomous driving research. Wang\textit{ et al.}\cite{wang2022social} emphasizes the mutual influence between interaction subjects, defining road traffic interaction as a dynamic behavioral sequence between two or more subjects to maximize individual benefits and minimize costs through the process of information exchange. Serving the purpose of sociality-aware autonomous driving, Zhao\textit{ et al.}\cite{zhao2024measuring} puts interaction as a situation where more than one motorized road user compromises part of its potential individual rewards as a result of taking into account group rewards. By defining interaction from the perspective of human decision-making, these studies provide a mathematical description of the dynamic processes of interaction. However, building a decision-making model that combines an optimization process could be challenging itself and subjective in detailed settings, for example, the reward terms. Therefore, such definitions still fall short of clarity and general applicability. 

While an increasingly concrete concept of driving interaction after years of effort, we still strive to identify driving interaction events from raw natural driving records with a cross-dataset and cross-research-applicable view. One of the possible cause is the semantic definitions do not provide the clarity needed to determine the precise timing and participants of driving interactions. Therefore, on the basis of \textit{what is interaction}, issues still need to be addressed to reach a quantified consensus on \textit{when and where is the interaction}.


\textit{Quantitative measurement of driving interaction}. In the quest for the practical needs of interaction research, methods for identifying and measuring the driving interaction have been developed. Rule-based methods extract interactions through a set of predefined rules, focusing on typical scenarios such as unprotected left turns, lane changes, merges, etc. For example, Arbabi\textit{ et al.} \cite{arbabi2022planning} extract approximately 6,000 lane-change segments from the NGSIM dataset by setting rules to determine the start and end time points of lane changes. Similarly, Liu\textit{ et al.}\cite{liu2022interaction} identify interacting vehicles in the merging scenarios using a selection box, with the front boundary set at a 2-second time headway in front of the merging ramp. As a relatively more general extension, Li\textit{ et al.} \cite{li2023comparative} develop heuristic rules to capture interaction events between two agents with crossed paths from the Argoverse-2 dataset. Rule-based methods, while effective at capturing desired interaction events in specific scenarios, often face challenges with both precision and generalizability. The inherent ambiguity in rule descriptions can lead to inconsistent interpretations and therefore outcomes. For example, the "selection box" introduced in Liu\textit{ et al.}'s work \cite{liu2022interaction} remains unclear in determining if a vehicle is "within the box" in cases of partial overlap. Such methods, designed for predefined scenarios, limit their applicability to diverse real-world conditions.

Metric-based methods step further into more generic interaction extraction, by calculating metrics related to vehicle motion and then selecting interaction segments that fall within predefined metric ranges. In this way, the subjective nature of semantic interpretation is replaced by measurable criteria, enhancing clarity and consistency in interaction detection. With fewer limitations to specific functional scenarios, metric-based methods provide a broader and more flexible framework for identifying interactions. Li\textit{ et al.} \cite{li2019interaction}, for instance, utilized a predefined distance threshold of 20 feet to identify interacting counterparts of a target vehicle. Shu\textit{ et al.} \cite{shu2023human} determined whether a vehicle was involved in an interaction by assessing if it exhibited an emergency deceleration greater than 2.5 m/s². Wilson\textit{ et al.} \cite{wilson2023argoverse}, on the other hand, measured participant density in a given area to assess the interestingness of trajectories within scenes. This approach pinpoints some multi-object interactions, while it might miss interactive scenarios or misclassify non-interaction scenes, as close trajectories are not necessarily interacting and intensive interaction could happen in scenes with few participants. Pioneering a new direction, Zhan\textit{ et al. }\cite{zhan2019interaction}introduced the minimum time-to-conflict-point difference (MTTCP) as a quantitative metric to capture interactions between two agents. This method was instrumental in developing the renowned INTERACTION dataset and has since become a standard in interaction extraction. Nevertheless, interactions are not confined to pairs of vehicles. They usually occur in multi-agent systems with interconnected actions, which are difficult to decouple into pairs. Consequently, the MTTCP method falls short when it comes to quantifying interactions involving complex chain reactions among multiple agents.

To overcome the limitations of existing approaches, a method to quantify and identify driving interactions is ideally precise in expression, unrestricted by agent number, and not confined to specific functional scenarios. By breaking down the semantic definition of driving interactions into core components that include interaction entities, spatio-temporal conflict, and mutual influence, we used formal language to define the driving interaction in a readable and rigorous way. An interaction quantifier is then developed to detect interaction events and measure their intensity without limitations on scenario type or agent number, facilitating the reliability and generalizability of the findings in driving-interaction-related fields.

\section*{Methods}
\label{sec:methods}

To handle multiple data resources, a data preprocessing interface is developed based on an open source toolkit \texttt{trajdata}\cite{ivanovic2024trajdata}, converting driving records into a unified format. Then, the formal method is employed to quantify and extract driving interaction events from driving data collections to form a dense driving interaction dataset. This section first introduces the data structure of the driving interaction events in the \texttt{InterHub}, followed by the detailed methodology for identifying interaction segments from raw driving records.

\subsection*{Overall data structure}

\begin{figure}
    \centering
    \includegraphics{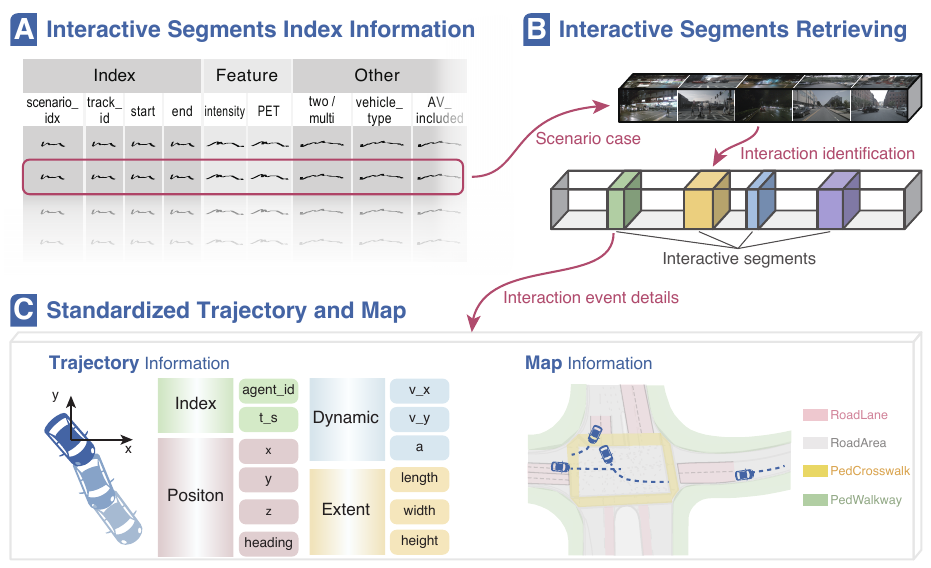}
    \caption{Overall data structure of InterHub}
    \label{fig:data_formats}
\end{figure}

As illustrated in Figure \ref{fig:data_formats}, \texttt{InterHub} is structured as a list, where each entry represents the metadata of an individual interaction scenario. This indexed list allows users to efficiently retrieve specific interaction events and access their detailed trajectory-level data as needed. The metadata for each interaction scenario is categorized into three types: indices, interaction features, and labels. Indices provide essential details for rapid identification and retrieval, including scenario indices, participant track IDs, and the start and end times of the interaction. Interaction features include metrics such as interaction intensity and PET (Post-Encroachment Time), reflecting the degree of interaction and potential conflict risk. Labels enable users to filter and interpret interaction scenarios based on specific criteria, such as the number of vehicles involved, vehicle types, and whether an autonomous vehicle participated in the interaction.


The interaction segments encompass both trajectory and map data, all in a consistent format. In the retrieved interaction segments, the trajectory of each vehicle is furnished at every time step along with the corresponding map data. The trajectory information encompasses three principal types: index (e.g. track ID, timestamp), position (e.g., x, y coordinates), heading, dynamics (e.g., speed, acceleration), and extent (length, width, height), as depicted in Figure \ref{fig:data_formats}C. In case dynamic data are absent from the original dataset, trajdata computes them through finite differences by default. The map elements processed by trajdata are categorized into four classes: RoadLane, RoadArea, PedCrosswalk, and PedWalkway. Users can also inquire about lane connectivity (i.e., lanes accessible from the left or right) and successor/predecessor lanes (i.e., lanes continuing from or leading to the current one).

\subsection*{Extracting interaction events using the formal method}

The diverse types and sparse distribution of driving interaction events, along with the lack of research on defining and quantifying their spatio-temporal scope, have made it challenging to comprehensively extract these events. In this section, we introduce a formalized method for describing and quantifying driving interactions.
Our approach builds on the widely adopted qualitative definition of driving interactions proposed by Markkula \textit{et al.} \cite{markkula2020defining}, which has been influential in driving interaction research \cite{azizi2023communication, cui2023passing, domeyer2022driver}. Markkula defines driving interaction as:

\begin{quote} \textit{A situation where the behaviour of at least two road users can be interpreted as being influenced by the possibility that they are both intending to occupy the same region of space at the same time in the near future.} \end{quote}

This definition establishes a unified concept for interactions among different types of road users, specifying three key elements:
\begin{itemize}
    \item Number of interaction entities: At least two road users are involved, with the possibility of multiple participants.
     \setlength{\itemsep}{0.001\baselineskip}
    \item Presence of spatiotemporal conflict: Interaction entities have the potential to occupy the same spatial area at the same time in the near future.
     \setlength{\itemsep}{0.001\baselineskip}
    \item Mutual influence: The spatiotemporal conflict prompts road users to take actions to avoid collisions, inevitably influencing one another's behavior.
\end{itemize}

\begin{figure}
    \centering
    \includegraphics{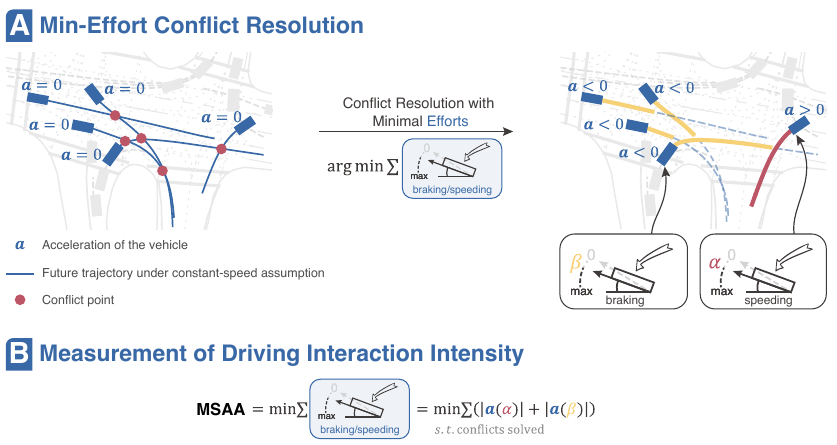}
    \caption{Interaction defined by min-effort conflict resolution.}
    \label{fig:interaction_definition}
\end{figure}

The semantic descriptions above pictured a readable scene of interaction, transferring the core sight of "interaction is where efforts have to be made for solving potential conflict".
Following this, we convert the semantic description using formal language, leveraging both the readability of semantics and the precision of mathematics\cite{maierhofer2022formalization}. The main idea is visualized in Figure \ref{fig:interaction_definition} and is mathematically detailed in the following text.

Given a driving scene, we identify the conflict point of any two of the vehicles as the intersections of their future path assuming their motion at current speed. Then, all vehicles interconnected through a chain of these conflicts are regarded as potentially involved in the interaction (left side of Figure \ref{fig:interaction_definition}A).

To navigate safely and orderly through conflict points, certain vehicles shall act to adjust their motion states. We consider their actions in the mean of longitudinal acceleration. Thus, given the initial scenario state, there is a minimal collective effort to resolve the conflict. The greater the minimum required change in motion state, the greater the impact of potential conflicts, leading to more intense interactions among the vehicles. And the effort is represented by the potential braking and acceleration actions taken by all vehicles (right side of Figure \ref{fig:interaction_definition}A). Therefore, the minimum sum of absolute acceleration or deceleration required to resolve potential conflicts (referred to as MSAA), is used to measure interaction intensity (Figure \ref{fig:interaction_definition}B).

\begin{figure}
    \centering
    \includegraphics[width=1\linewidth]{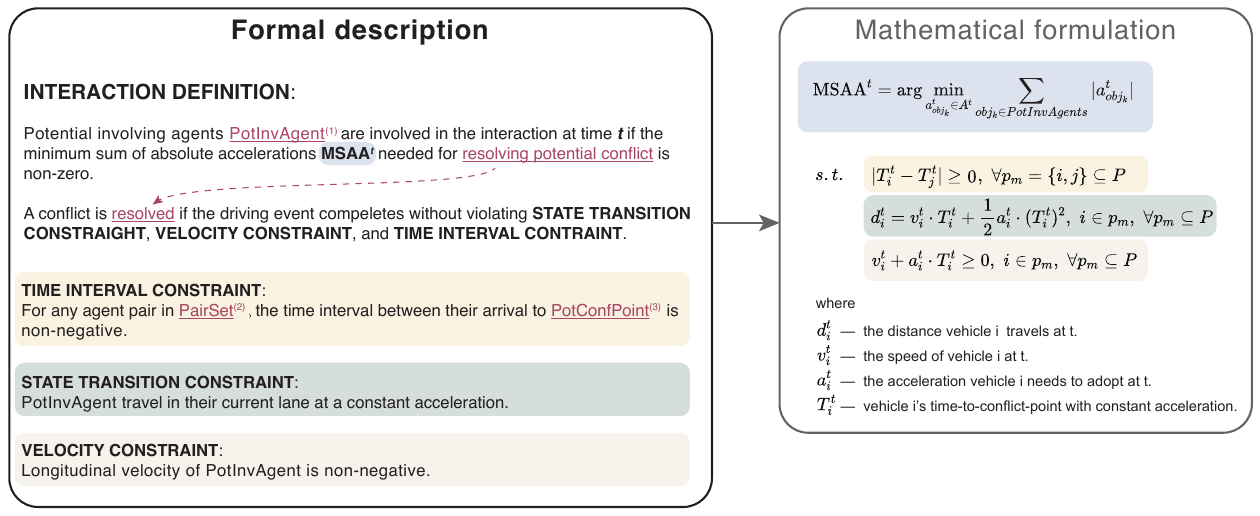}
    \caption{Formal description of interaction.}
    \label{fig:method}
\end{figure}

Our main ideas for defining interaction are emphasized in Figure \ref{fig:method}. MSAA\textsuperscript{\textit{t}}, derived by solving an optimization problem, is employed herein to ascertain whether an interaction occurs at timestep \textit{t}. The optimization problem uses the required acceleration or deceleration for each vehicle as variables, based on the vehicle's current motion state (including speed and distance to the conflict point) under the assumption of uniform motion. The feasibility of the current acceleration or deceleration in resolving the conflict is determined by applying the State Transition Constraint, Velocity Constraint, and Time Interval Constraint, as illustrated in the figure.

The above semantic idea is translated into a formal description using Metric Temporal Logic (MTL), whose formulas(such as $\phi$) are constructed from simple atomic propositions (a Boolean statement) and combined with logical operators(\( \land \) and \( \lor \) indicates the conjunction and disjunction, \(\phi_1 \Rightarrow \phi_2\) and \(\phi_1 \Leftrightarrow \phi_2\) represents logical implication and logical equivalence) and temporal operators(\(G_I(\phi)\) indicates that $\phi$ holds throughout the entire time sequence \(I \in \mathbb{R}^+ \), \(F_I(\phi)\) indicates that $\phi$ holds within the time interval  \(I \in \mathbb{R}^+ \) for some future states) to form more complex logical expressions. Additional key concepts are detailed as follows.

\noindent (1) \textbf{\textit{PotInvAgent:}} \textbf{The set of agents potentially involved in interaction} consists of all the vehicles(denoted as $\textit{obj}_1$, $\textit{obj}_2$, ...) that involving in the same \underline{ChainConflict \textsuperscript{(1.1)}}:
\begin{align*} 
\text{PotInvAgents} = \{\textit{obj}_1, \textit{obj}_2, \ldots, \textit{obj}_n\} \Longleftrightarrow \text{ChainConflict}(\textit{obj}_1, \textit{obj}_2, \ldots, \textit{obj}_n)
\end{align*}

\begin{quote}
(1.1) \textbf{\textit{ChainConflict : }}\textbf{A series of interconnected interactions between multiple agents} where \underline{STConflict\textsuperscript{(1.2)}} propagate through direct and indirect influences. It's defined as:
\begin{align*}
\text{ChainConflict}(\textit{obj}_{1}, \textit{obj}_{2}) &\Longleftrightarrow \text{STConflict}(\textit{obj}_{1}, \textit{obj}_{2}) \\  
\text{ChainConflict}(\textit{obj}_{1}, \textit{obj}_{2}, \ldots, \textit{obj}_{n}, \textit{obj}_{n+1}) &\Longleftarrow \text{ChainConflict}(\textit{obj}_{1}, \textit{obj}_{2}, \ldots, \textit{obj}_{n}) \wedge \\ 
& \phantom{\Longleftarrow C} \exists \textit{i} \in \left[1,n\right] \wedge \text{STConflict}(\textit{obj}_{n+1}, \textit{obj}_{i})
\end{align*}

(1.2) \textbf{\textit{STConflict}}: \textbf{Spatiotemporal conflict} implies circumstances in which two vehicles from different lanes are projected to pass through the intersection of predicted future trajectories almost simultaneously in the immediate future. Specifically, it means that there exists an \underline{Intersection\textsuperscript{(1.3)}} between the estimated future trajectories (calculated by \( \text{FuTraj}(\textit{obj}_{i}, m) \)) of the two agents over the next \(m\) (where \(m\) is set to 5 seconds) time steps, and the time difference when they pass through this intersection point is less than \( \textit{ConfTime} \) (a threshold set to 3 seconds), as determined by the function \( \text{TimeDiffLimit} \). The function, \( \text{Time2inter}\left(x\left(\textit{obj}_i\right), \text{IntersectPoint}\left(\textit{traj}_1, \textit{traj}_2\right)\right) \)
, calculates the time required for \( \textit{obj}_i \) to reach the \( \text{IntersectPoint} \) (a function that calculates the intersection point of two trajectories) following its current state \( x(\textit{obj}_i) \). 
\begin{align*}
\text{STConflict}(\textit{obj}_{1}, \textit{obj}_{2}) \Longleftarrow & \text{Intersection}\left(\text{FuTraj}\left(\textit{obj}_{1}, m\right), \text{FuTraj}\left(\textit{obj}_{2}, m\right)\right) \wedge \\
& \text{TimeDiffLimit}\left(\text{Time2inter}\left(x\left(\textit{obj}_{1}\right), \text{IntersectPoint}\left(\textit{traj}_{1}, \textit{traj}_{2}\right)\right), \right. \\
& \phantom{\text{TimeDiffLimit}(} \left. \text{Time2inter}\left(x\left(\textit{obj}_{2}\right), \text{IntersectPoint}\left(\textit{traj}_{1}, \textit{traj}_{2}\right)\right), \textit{ConfTime}\right)
\end{align*}

(1.3) \textbf{\textit{Intersection}}: The logic for determining whether two trajectories intersect is as follows, where \( \text{BufferLine}(\textit{traj}_i, n) \) calculates the \( n \)-meter buffer of the trajectory, \( \text{BufferPolygon}(\textit{traj}_i, n) \) calculates the \( n \)-meter buffer polygon of the trajectory, and Points is the set of all coordinate points in the current scene. The function \( \text{Outside}(\textit{Line}, \textit{Polygon}) \) checks whether the given line starts or moves from outside of the specified polygon.
\begin{align*}
\text{Intersection}\left(\textit{traj}_{1}, \textit{traj}_{2}\right) \Longleftarrow & \text{IntersectPoint}\left(\textit{traj}_{1}, \textit{traj}_{2}\right) \in \text{Points} \wedge \\
& \phantom{(}\text{IntersectPoint}\left(\textit{traj}_{2}, \text{BufferLine}\left(\textit{traj}_{1}, n\right)\right) \in \text{Points} \wedge \\
& \phantom{(}\text{Outside}\left(\textit{traj}_{2}, \text{BufferPolygon}\left(\textit{traj}_{1}, n\right)\right)
\end{align*}

\end{quote}

\noindent (2) \textit{\textbf{PairSet: }}The set of all agent pairs that share the same potential conflict points:
\begin{align*}
\text{PairSet} = \{ (obj_i, obj_j) \mid obj_i \in \text{PotInvAgents}, obj_j \in \text{PotInvAgents}, \text{STConflict}(obj_i, obj_j) \}
\end{align*}

\noindent (3) \textbf{\textit{PotConfPoint:}} Intersection point of two trajectories if there's spatiotemporal conflict:
\begin{align*}
\text{PotConfPoint}\left(\textit{obj}_{1}, \textit{obj}_{2}\right) = \text{IntersectPoint}\left(\text{FuTraj}\left(\textit{obj}_{1}, m\right), \text{FuTraj}\left(\textit{obj}_{2}, m\right)\right) \Longleftarrow \text{STConflict}\left(\textit{obj}_{1}, \textit{obj}_{2}\right)
\end{align*}

\subsubsection*{Complete interactive driving segment}

To filter the complete interactive driving segment, the status of the set of involved agents (InvolvedAgents) should adhere to the following logic:
\begin{align*}
(\text{InvolvedAgents} \;&= \text{PotInvAgents})\wedge (\text{SegmentStart} \;= f_{0}) \wedge (\text{SegmentEnd} \;= f_{\text{end}}) \\
&\Longleftarrow (G_{[f_{0}]}(\text{IntCheck}(\text{PotInvAgents})) \wedge G_{[f_{\text{end}}]}(\text{IntCheck}(\text{PotInvAgents}))) \wedge \\
&\quad \quad(G_{[f_{1}, f_{2}]} (\neg \text{IntCheck}(\text{PotInvAgents})) \wedge [f_{1}, f_{2}] \subseteq [f_{0}, f_{\text{end}}] \rightarrow \\
&\quad \quad \; F_{[f_{1}-3, f_{1}]} (\text{IntCheck}(\text{PotInvAgents})) \wedge F_{[f_{2}, f_{2}+3]} (\text{IntCheck}(\text{PotInvAgents})))
\end{align*}
where [$f_{0}$, $f_{end}$] represents the time range of the interaction event, IntCheck(PotInvAgents) represents if the condition that MSAA > threshold is satisfied. This defines the involved agents (\text{InvolvedAgents}) and the start (\text{SegmentStart}) and end (\text{SegmentEnd}) times  of the interaction segment.

\section*{Data records}
\texttt{InterHub} provides both a dense driving interaction dataset and a dedicated toolkit to expand the dataset using user-hand data resources be it either public or private. Extracted interaction events data are freely accessible at \href{https://figshare.com/articles/dataset/_b_InterHub_A_Naturalistic_Trajectory_Dataset_with_Dense_Interaction_for_Autonomous_Driving_b_/27899754}{figshare repository}. The raw data for the original datasets can be found in the corresponding literature or websites which are also noted in the documents. The toolkit for expanding the dataset is available at github repository \href{https://github.com/zxc-tju/InterHub}{https://github.com/zxc-tju/InterHub}.

\section*{Technical validation}
We validate the extracted and unified driving interaction events from two aspects. First, the interaction events are analyzed through number of involved agents, interaction duration, interaction intensity, and safety. Second, we demonstrate the value of the proposed dense interaction dataset in advancing autonomous driving by challenging the state-of-the-art solutions in two key autonomous driving tasks: motion prediction and motion planning.

\subsection*{Characteristics of driving interaction events}

Following the interaction defined by min-effort conflict resolution, we extracted interaction events from the nuPlan, Waymo Open Motion, Lyft Level 5, and INTERACTION dataset, respectively. Statistical data are presented in Table \ref{tab:event_stat}. Detailed characteristics of the interaction events from these datasets are discussed.

\begin{table}[h]
    \centering
    \begin{tabular}{|c|c|c|c|c|c|}
        \hline
        \textbf{Dataset} & \textbf{Num. of interaction events} & \textbf{Intensity (m/s²)} & \textbf{Num. of agents} & \textbf{Duration (s)}  & \textbf{Min. PET (s)} \\
        \hline
        Waymo   & 28,666   & 2.02   & 64,256   & 1.29   & 0.84   \\
        nuPlan  & 10,296  & 1.30   & 22,350   & 1.14   & 0.85   \\
        Lyft    & 7,017   & 1.67   & 15,013    & 1.22   & 0.77   \\
        INTERACTION & 19,420 & 2.04  & 44,551    & 0.88   & 0.67   \\
        \textbf{InterHub}  & 65,399  & 1.76     & 146,170    & 1.13     & 0.79     \\
        \hline
    \end{tabular}
    \caption{Statistics of driving interaction events between \texttt{InterHub} and raw datasets}
    \label{tab:event_stat}
\end{table}

\subsubsection*{Typical interaction cases}
\begin{figure}
    \centering
    \includegraphics[width=1\linewidth]{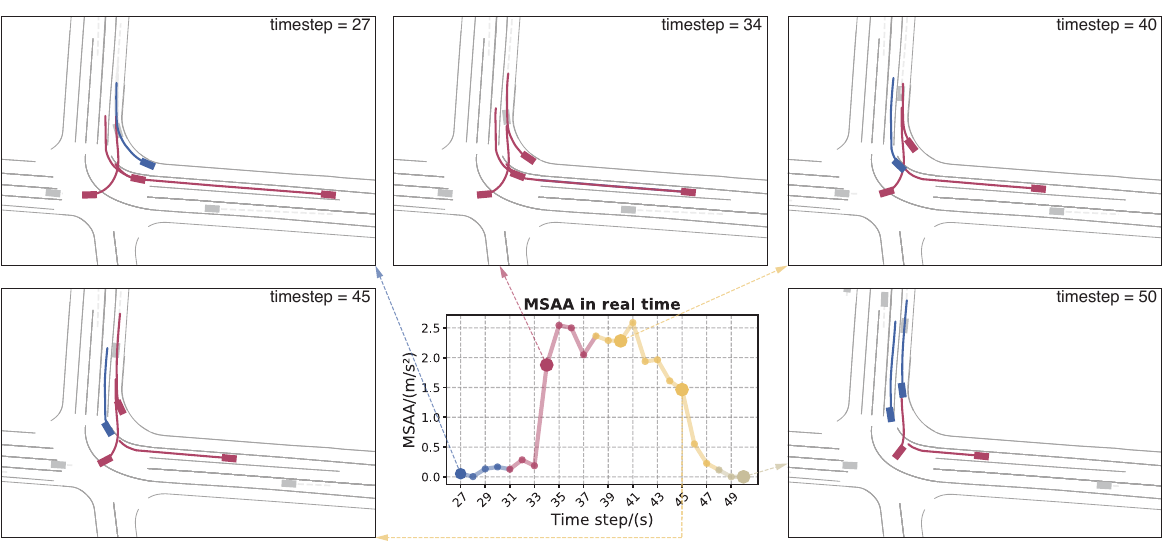}
    \caption{Dynamic process of a driving interaction case. Interacting agents (with most efforts assigned to them in resolving conflicts) are highlighted in red, while other agents are marked in blue.}
    \label{fig:scenario-case}
\end{figure}

Figure \ref{fig:scenario-case} illustrates the process of a driving interaction case among multiple vehicles, derived from the Waymo Open Motion dataset. It depicts the dynamics of vehicle interactions at an intersection and the shifts of key interacting agents engaging in interaction. The solid lines in the figure denote the future trajectories of the vehicles over the subsequent 3 seconds. Different colors in the middle line chart represent different objects participating in the interaction. 

The scenario start when \#1 and \#2 encounter at the junction, with \#3 following \#2, resulting an interaction intensity of 0.1 m/s² which indicates a 0.1 m/s² acceleration for collision avoidance.
By timestep 34, with the involvement of \#4, the interaction intensity increases to around 2 m/s². \#2 successfully passes through its conflict point with \#1 at timestep 37, leading to the shift of interaction objects from all vehicles to \#1, \#3 and \#4. From timestep 41, the interaction intensity is decreasing, possibly because some vehicles have already taken measures to slow down. At timestep 48, when \#4 reaches the conflict point, the interactive agents shift to \#1 and \#3. By timestep 50, the interaction ends when \#1 reaches its point of conflict with \#3.

\begin{figure}
    \centering
    \includegraphics[width=1\linewidth]{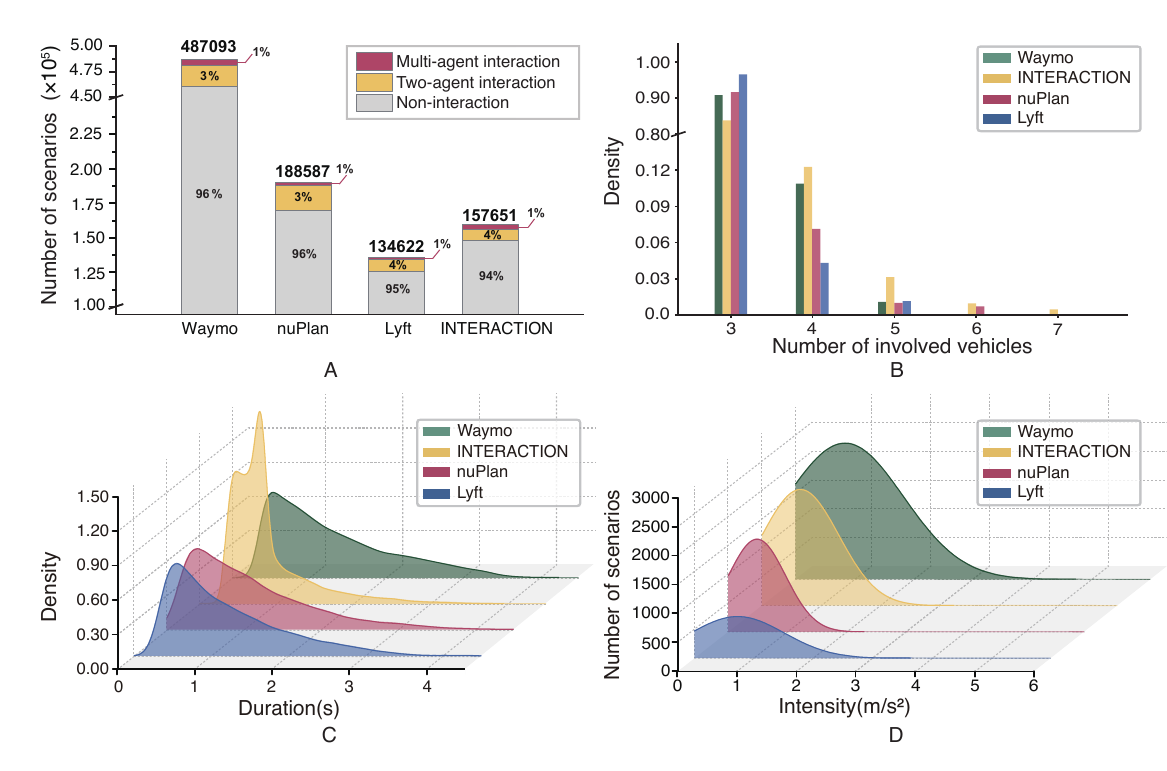}
    \caption{Results of interaction extraction: (A) quantity of scenarios, (B) density of number of vehicles participating in multi-agent interaction events, (C) distribution of effective interaction duration of the driving interaction segments, (D) distribution of interaction intensity of the driving interaction segments.}
    \label{fig:extract-result}
\end{figure}

\subsubsection*{Interaction distribution}

As illustrated in Figure \ref{fig:extract-result}A, although raw naturalistic driving datasets encompass a substantial number of scenarios, the proportion of those that contain valid interaction events is relatively low. "Two-agent interaction scenarios" refer to instances that exclusively involve interactions between two vehicles, whereas "multi-agent interaction scenarios" denote situations where interactions among multiple vehicles occur. The proportion of two-agent interaction scenarios, which is approximately 3\%, and the proportion of multi-agent interaction scenarios, estimated at around 1\%, remain consistent across all datasets,which shows that the distribution of driving interaction scenarios in existing datasets is extremely sparse.

\subsubsection*{Interaction participants}

Figure \ref{fig:extract-result}B depicts the density distribution of vehicles engaged in multi-agent interaction events. It is noteworthy that the nuPlan and INTERACTION dataset exhibit higher proportions of multi-vehicle interaction scenarios involving a greater number of participating vehicles. On average, multi-vehicle interaction scenarios in the INTERACTION dataset comprises 3.2 vehicles, with the maximum value being 7. As the number of participants can serve as an indirect indicator of the complexity inherent in these scenarios, the INTERACTION dataset may feature more complex interaction events compared to other datasets. The maximum number of interacting vehicles within the Waymo and Lyft datasets amounts to merely 5, which implies that the complex multi-vehicle scenarios might be inadequately represented.

\subsubsection*{Interaction duration}

Figure \ref{fig:extract-result}C illustrates that across all datasets, the effective interactions are temporally sparse. The duration of interactions is predominantly concentrated in the range of 0.25 to 3.5 seconds within the extracted segments with maximum durations seldom exceeding 4 seconds. The distribution of effective interaction durations across the Waymo and Lyft datasets is similar, with a majority of interactions concentrated around 0.6 seconds. Compared to the above two datasets, the nuPlan dataset distribution is slightly skewed to the left and the maximum distribution is around 0.4s. It is worth noting that since the original datasets are divided into 4 seconds fragments, the average interaction duration of the INTERACTION dataset is shorter. Moreover, its distribution is quite different from the other datasets, which have two peaks, with one around 0.4 seconds and the other around 0.75 seconds.

\subsubsection*{Interaction intensity}
Figure \ref{fig:extract-result}D illustrates the distribution of interaction intensity across driving segments derived from each datasets. Leveraging its extensive database, the Waymo dataset yields the highest number of interaction segments. Conversely, the fewest interaction fragments were extracted from the Lyft dataset. Meanwhile, the interaction segments provided in the nuPlan, Lyft, and INTERACTION datasets are predominantly concentrated within a range where interaction intensity is below 2.3 m/s². Only a limited number of segments demonstrate stronger interactions. Compared to other datasets, the Waymo dataset contains a higher number of scenarios with interaction intensity in the range of 2.3 m/s² to 4 m/s². This suggests that, due to the larger base volume of data in Waymo, even relatively rare scenarios with higher interaction intensity are more likely to be present. 



\subsubsection*{Safety assessment}

\begin{figure}
    \centering
    \includegraphics[width=1\linewidth]{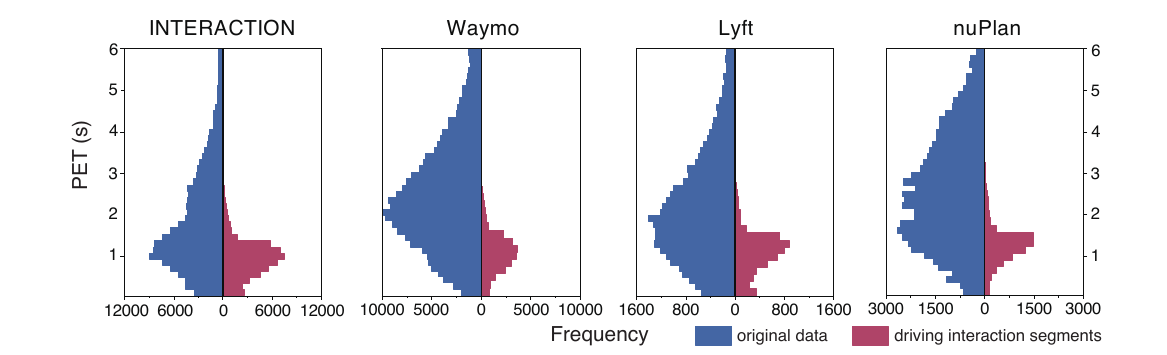}
    \caption{Distributions of PET for different segments of driving interaction.}
    \label{fig:pet}
\end{figure}

While interaction intensity offers a comprehensive, and typically beyond-safety, perspective on the strength of interactions, it stands align with the safety-oriented interaction definition. Given that Post-Encroachment Time (PET), a surrogate safety metric, has a clear and distinct physical interpretation, it serves as an effective indicator for assessing the safety of driving interaction events. PET denotes the temporal interval between an encroaching road agent exiting the conflict zone and another road agent entering that same zone. We compare minimum PET distributions of driving interaction segments for four datasets. The results are presented in Figure \ref{fig:pet}.
Note that we did not use any safety index to filter in the interaction cases, but the dangerous cases indicated by PET are still extracted. A typical example of those dangerous-but-not-interactive cases is "a following car gets close to its front car at a high relative speed".

\subsection*{Versatile benchmark for key autonomous driving tasks}

Autonomous driving tasks, such as motion prediction and planning, are typically validated in naturalistic driving datasets. However, due to the sparsity of challenging scenarios occurring in the natural collection, the performance of corresponding solutions in critical scenarios can be obscured by the predominance of less interesting cases. This imbalance can lead to deceptively promising statistical measures. In this subsection, we reveal the performance degradation of state-of-the-art motion predictors and planners when transitioning from naturalistic driving datasets to the dense interaction subset of driving scenarios, highlighting the value of \texttt{InterHub} in providing a versatile benchmark for ensuring robust safety guarantees of AVs.

\subsubsection*{Motion prediction}

To validate the performance of autonomous driving tasks on our dense driving interaction dataset, baseline prediction models for each dataset are selected. We investigate the performance of three highly effective approaches for trajectory prediction based on the INTERACTION, Waymo and Lyft datasets. Given that the nuPlan dataset serves as a benchmark for planning in autonomous driving, it is predominantly utilized within the domain of motion planning. Consequently, an interactive prediction and planning framework representing a cutting-edge approach on the nuPlan benchmark is selected, and its prediction results are evaluated in this section. The following presents more detailed information of the baseline models:

\begin{itemize}

    \item \textbf{UQnet}\hspace{0.5em}\cite{li2024unravelling}, our previous work, attained leading performance in the INTERPRET Challenge, ranking first on the leader-board of Single-Agent Track\footnote{INTERPRET: Interaction-dataset-based prediction challenge, single-agent track, organized by ICCV 2021 Competition. Available at: \url{https://challenge.interaction-dataset.com/leader-board/}. Last accessed on September 4-th 2024.}. Using a deep learning model based on two-dimensional histograms and incorporating deep ensemble techniques, UQnet is capable of quantifying both epistemic uncertainty and aleatoric uncertainty, thereby significantly enhancing the generalizability of the missing rate compared to its previous state-of-the-art.
    
    \setlength{\itemsep}{0.001\baselineskip}
    \item \textbf{MTR}\hspace{0.5em}\cite{shi2022motion} is an innovative multimodal motion prediction framework that has achieved the top ranking on the Waymo Open Motion Dataset challenge leader-board\footnote{Waymo Open Dataset: Motion Prediction Challenge 2022, organized by Waymo. Available at: \url{https://waymo.com/open/challenges/2022/motion-prediction/}.}. MTR models motion prediction as a joint optimization of global intent positioning and local motion refinement. Experimental results on the Waymo Open Motion Dataset demonstrate that MTR achieves state-of-the-art performance in both edge and joint motion prediction tasks.
    
    \setlength{\itemsep}{0.001\baselineskip}
    \item \textbf{ContextVAE}\hspace{0.5em}\cite{xu2023context} is a context-aware model designed for predicting vehicle trajectories, which integrates environmental and agent dynamics through dual attention mechanisms to produce accurate, socially aware, and map-compliant predictions in real time. Since the lyft dataset did not have a unified race leaderboard to refer to, ContextVAE is selected as the baseline model due to its leading performance on lyft level 5 in current publications\cite{vazquez2022deep, salzmann2020trajectron}. 
    
    \setlength{\itemsep}{0.001\baselineskip}
    \item \textbf{GameFormer}\hspace{0.5em}\cite{huang2023gameformer} is a planning framework that builds upon a Transformer-based neural network that specifically designed for interactive prediction and planning. It also demonstrated leading performance in interaction prediction and planning in the final leaderboard of the NuPlan Challenge on the private test set\footnote{NuPlan Planning Challenge, 2023, organized by Motional. Available at :\url{https://eval.ai/web/challenges/challenge-page/1856/leaderboard/4360}.}. 
\end{itemize}

To comprehensively assess the performance of these models, the following metrics are applied. For each scenario, the model generates \( K \) joint predictions \( \hat{\mathbf{S}}^{(k)} = \{ \hat{\mathbf{s}}_{i,t}^{(k)}, i \in (1, \dots, N), t \in (1, \dots, T) \}, k \in (1, \dots, K) \), where predictions are made for \( T \) future time steps for \( N \) agents. In a similar manner, the ground truth is represented as \( \mathbf{S} = \{ \mathbf{s}_{i,t} \} \).

\begin{itemize}
    \item \textbf{minADE} \hspace{0.5em} The mean of the L2 norm between the real trajectory and the predicted trajectory for all agents over all time steps is as follows:
        \begin{equation}
        \text{minADE} = \min_{k} \frac{1}{NT} \sum_{i=1}^{N} \sum_{t=1}^{T} \| s_{i,t} - \hat{s}_{i,t}^{(k)} \|_2
        \end{equation}

    \setlength{\itemsep}{0.001\baselineskip}
    \item \textbf{minFDE} \hspace{0.5em} The L2 norm between the real trajectory and the predicted trajectory for all agents at the final predicted time step is as follows:
        \begin{equation}
        \text{minFDE} = \min_{k} \frac{1}{N} \sum_{i=1}^{N}  \| s_{i,T} - \hat{s}_{i,T}^{(k)} \|_2
        \end{equation}

    \setlength{\itemsep}{0.001\baselineskip}
    \item \textbf{Miss Rate (MR)} \hspace{0.5em} The proportion of predicted trajectories that deviate beyond a predefined threshold from the actual trajectory at the final predicted time step is as follows:
    \begin{equation}
        \text{MR} = \frac{1}{N} \sum_{i=1}^{N} \mathbb{I} \left( \min_{k} \| s_{i,T} - \hat{s}_{i,T}^{(k)} \|_2 < \delta \right)
    \end{equation}
    where $\mathbb{I}$ is an indicator function, when the minimum distance between the predicted value $\hat{s}_{i,T}^{(k)}$  and the ground truth value $s_{i,T}$ is smaller than a threshold $\delta$, the indicator function returns 1; otherwise, it returns 0.

\end{itemize}


\textit{Implementation details.}\hspace{0.5em} Each baseline model is trained on the training set, validated on the validation set, and tested on both the full test set and the test set that contains only interactive segments.  The training, validation, and full test sets are provided by the corresponding dataset of each model, whereas the test set containing only interactive segments consists of the 1000 segments with the highest interaction intensity from \texttt{InterHub}, the dense driving interaction dataset developed in this study. 

\textit{Model performance.}\hspace{0.5em} The corresponding results are presented in Table \ref{tab:model_performance}  . In contrast to the full testing datasets, significant performance variations are observed when employing interaction-only testing datasets. With interaction-only testing datasets, minADE, minFDE, and MR exhibit notable increments, indicating the inferior performance of these baseline models in prediction tasks in highly interactive scenarios. 
These results imply that the highly interactive datasets established in this study can offer more valuable challenges to autonomous driving tasks related to motion prediction.

\begin{table}[ht]
\centering
\begin{tabular}{|l|l|c|c|c|}
\hline
\textbf{Model} & \textbf{Dataset} & \textbf{MinADE(m)} $\downarrow$ & \textbf{MinFDE(m)} $\downarrow$ & \textbf{MR} $\downarrow$ \\
\hline
\multirow{2}{*}{\textbf{UQnet}} & Full (INTERACTION) & 1.01 & 0.66 & 0.60 \\
 & InterHub & 1.36 & 0.74 & 1.65 \\
 & \textit{Performance degradation} & \textit{34.65\%} & \textit{12.12\%} & \textit{175\%} \\
\hline
\multirow{2}{*}{\textbf{MTR}} & Full (Waymo) & 0.75 & 1.54 & 0.19 \\
 & InterHub & 0.92 & 1.78 & 0.36 \\
 & \textit{Performance degradation} & \textit{22.67\%} & \textit{15.58\%} & \textit{89.47\%} \\
\hline
\multirow{2}{*}{\textbf{ContextVAE}} & Full (Lyft) & 0.24 & 0.54 & - \\
 & InterHub & 0.5869 & 1.2776 & - \\
 & \textit{Performance degradation} & \textit{144.54\%} & \textit{136.59\%} & \textit{-} \\
\hline  
\multirow{2}{*}{\textbf{Gameformer}} & Full (nuPlan) & 0.9306 & 1.8525 & - \\
 & InterHub & 1.2603 & 2.9247 & - \\
 & \textit{Performance degradation} & \textit{35.43\%} & \textit{57.88\%} & \textit{-} \\
\hline
\end{tabular}
\caption{Degraded performance of baseline models on the dense interaction dataset \texttt{InterHub}. The degradation in performance is significant for all models at the confidence level of $\alpha=0.05$.}
\label{tab:model_performance}
\end{table}

\subsubsection*{Motion planning}

In this subsection, we present the results of the planning and control tasks within challenging interaction scenarios. These tasks were evaluated as part of the OnSite Autonomous Driving Algorithm Challenge (\url{www.onsite.com.cn}) held from 2023 to 2024. Organized by Tongji University and co-hosted by the China National Intelligent Connected Vehicle Innovation Center, the challenge attracted over 300 teams from more than 70 universities and institutions worldwide. With a specific focus on interaction scenarios such as head-on, crossing path, and merging path cases, selective scenarios were incorporated into the competition to challenge candidate planners. 

 We evaluated the average performance scores of the top 10 planners across over 700 interaction scenarios from the challenge leaderboard. Their task records were analyzed to investigate the relationship between algorithm performance and the interaction intensity of the scenarios. Performance scores were measured based on three primary factors: safety (time to collision, collision occurrence, and road boundary violations), efficiency (task completion and duration), and comfort (longitudinal and lateral acceleration, jerk, and yaw rate),  using the \href{https://www.onsite.com.cn/#/dist/evaluationTool} {evaluation tool} provided by the OnSite challenge. Interaction intensity in each scenario was quantified by the average \( \text{MSAA}^t \) over the involved agents during the interaction period.

\begin{figure}
    \centering
    \includegraphics[width=1\linewidth]{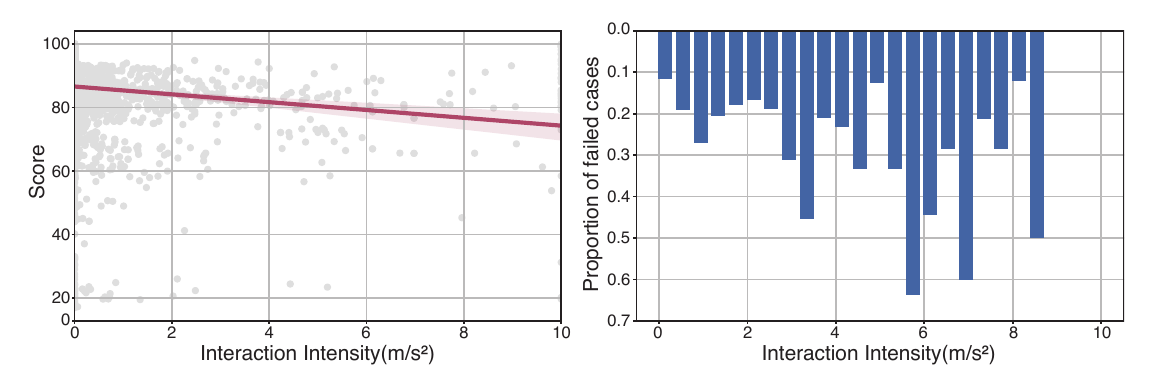}
    \caption{Planning scores over varied interaction intensity.}
    \label{fig:Average Scores}
\end{figure}

As shown in Figure \ref{fig:Average Scores}, a consistent pattern emerges: as interaction intensity increases, the average performance scores of the algorithms tend to decrease, and the proportion of algorithms that fail to complete the task rises. This indicates that higher interaction intensity introduces escalating challenges for planning algorithms.

Mining long-tailed, high-risk scenarios is crucial for testing autonomous vehicles. Our findings suggest that verifying AV performance under such highly interactive conditions is essential for addressing the "tail" of challenging cases, which are critical for ensuring robust safety guarantees. To this end, \texttt{InterHub} is proposed not only as a dataset containing dense interaction cases but also as a hub for identifying and quantifying interaction events in a manner that is universally applicable across various disciplines.

\section*{Code availability}

The codes for the unification of the raw driving records and the extraction and analysis of interaction events are documented and publicly available at \href{https://github.com/zxc-tju/InterHub}{https://github.com/zxc-tju/InterHub}. We provide three tools to help users navigate \texttt{InterHub}:
\begin{itemize}
  \item \textit{0\_data\_unify.py} provides a data interface to convert various data resources into a unified format that works seamlessly with the interaction event extraction process.
  \item \textit{1\_interaction\_extract.py} extracts interactive segments from the unified driving records, following the criterion detailed in \hyperref[sec:methods]{Methods}.
  \item \textit{2\_case\_visualize.py} showcases the process of retrieving and visualizing typical interaction cases.
\end{itemize}

Codes and data usage is restricted to research purposes only. Any commercial exploitation of the data requires separate approval and possibly additional agreements.

\section*{Acknowledgements}

This research is jointly sponsored by National Natural Science Foundation of China (52125208, 52232015) and National Key R\&D Programs of China (2023YFB4301900).

\section*{Author contributions}

X.J. and X.Z. coded for the data extraction, X.Z., Y.L., and Z.L. reviewed the codes, J.S. and X.Z. conceived the experiments, X.J., Y.L., and Z.L. conducted the experiments, X.J. and X.Z. analyzed the results. All authors drafted the manuscript and reviewed the data. 

\section*{Competing interests}

The authors declare no competing interests.

\end{document}